\UseRawInputEncoding
\documentclass[3p,twocolumn]{elsarticle}

\usepackage{amssymb}

\usepackage{amsmath}
\usepackage{mathtools}

\usepackage{changepage}
\usepackage{tabularx}
\usepackage{bm}  
\usepackage{textcomp}
\usepackage{natbib}
\usepackage{hyperref}

\journal{Mechanics Research Communications}

\newcommand{\ubar}[1]{\mkern 1.5mu\underline{\mkern-1.5mu#1\mkern-1.5mu}\mkern 1.5mu}
\newcommand{\uubar}[1]{\ubar{\ubar{#1}}}

\begin{document}

\begin{frontmatter}


\title{Neural-network acceleration of projection-based model-order-reduction for finite plasticity: Application to RVEs}
\author[add1,add2]{S. Vijayaraghavan}
\author[add2]{L. Wu}
\author[add2]{L. Noels}
\author[add1]{S.P.A. Bordas}
\author[add3]{S. Natarajan}
\author[add1]{L.A.A. Beex\corref{cor1}}
\ead{lars.beex@uni.lu}
\cortext[cor1]{Corresponding author}
  \address[add1]{University of Luxembourg, Faculty of Science, Technology and Medicine: 6 Avenue de la Fonte, Esch Sur Alzette, Luxembourg,\href{https://legato-team.eu}{Legato-Team} }
\address[add2]{University of Liege, Bвt. B52/3 Computational \& Multiscale Mechanics of Materials, Quartier Polytech 1, allйe de la Dйcouverte 9 ,4000 Li\`ege, Belgium}
\address[add3]{Department of Mechanical Engineering, Indian Institute of Technology, Madras, Chennai - 600036, India.}




\begin{abstract}

Compared to conventional projection-based model-order-reduction, its neural-network acceleration has the advantage that the online simulations are equation-free, meaning that no system of equations needs to be solved iteratively. Consequently, no stiffness matrix needs to be constructed and the stress update needs to be computed only once per increment. In this contribution, a recurrent neural network is developed to accelerate a projection-based model-order-reduction of the elastoplastic mechanical behaviour of an RVE. In contrast to a neural network that merely emulates the relation between the macroscopic deformation (path) and the macroscopic stress, the neural network acceleration of projection-based model-order-reduction  preserves all microstructural information, at the price of computing this information once per increment.

\end{abstract}




\end{frontmatter}


\section{Introduction}

Relatively recently, artificial neural networks (ANNs) have been investigated to emulate the relation between the macroscale deformation (path) and the macroscale stress in nested multiscale approaches \cite{GHAVAMIAN2019112594, SETTGAST2020102624, UNGER20081994, 10.3389/fmats.2019.00075, LOGARZO2021113482}. Although such ANN-emulations are rapid, all microstructural information is in principle lost (some microstructural information could be included in the ANN emulator \cite{inpreparation}). In order to preserve all microstructural information, ANNs can be combined with projection-based model-order-reduction (MOR) \cite{KORONAKI2019148,SWISCHUK2019704}.

Projection-based MOR is an \textit{a posteriori} method; it utilizes the solutions of training simulations as global basis functions to interpolate kinematic variables. It uses either a representative set of orthonormalized training solutions directly as global basis (i.e.~the method of `reduced basis' \cite{rb1,rb2}), or applies singular value decomposition to the training solutions, and uses the basis vectors associated with the highest singular values as global basis functions (i.e.~the method of `Proper Orthogonal Decomposition' - POD \cite{Kerfriden2013,Kerfriden2011a,Meyer2003,nme3050}).

The global basis in projection-based MOR interpolates the kinematic variables to reduce the number of degrees of freedom in the online simulations. In this contribution, ANNs are used to emulate the values of these remaining degrees of freedom: the coefficients of the global basis functions. This eliminates the need to construct stiffness matrices, since the iterative process to solve for the basis coefficients is avoided. The only issue that remains to be computed once per increment is the stress update in each quadrature point (i.e.~the plastic variables in the case of elastoplasticity).

The aim of this work is to formulate an ANN-accelerated POD-based MOR for finite plasticity under cyclic and random loading, applied to a representative volume element (RVE). The use of projection-based MOR for non-elliptical problems (such as those governed by elastoplasticity) requires a large number of global basis functions to achieve an acceptable accuracy (we use 100 basis functions). As ANNs avoid the computation of the basis coefficients, many basis functions can be used and hence, ANN-accelerated projection-based MOR may be considered particularly useful in the context of non-elliptical problems. 

Since elastoplasticity includes both reversible and irreversible physics, a suitable ANN must be able to account for the deformation path. Since \cite{Mozaffar26414,GORJI2020103972,WU2020113234,GHAVAMIAN2019112594} have shown that the hidden variables in recurrent neural networks (RNNs) are able to account for this (in the context of conventional finite element simulations to compute inelastic responses), these types of ANNs are also used in the current contribution.

The remainder of this paper is organized as follows: in the next section, the direct numerical simulations are concisely discussed. Section 3 describes a conventional POD-based MOR followed by the description of the neural network architecture in section 4. Results are discussed in section 5, where the predictions of the RNNs are compared with those of the direct numerical simulation (DNS) and the conventional POD-based MOR. A short conclusion is presented in section 6.

\section{Direct Numerical Simulations}

The plane strain simulations employ bilinear quadrilateral (four node) finite elements with four Gauss quadrature points. An F-bar method is utilized to alleviate locking due to the incompressibility of the plastic deformation. Within this framework, the volume change of the deformation gradient tensor at a quadrature point is replaced with the volume change at the center of the element. The resulting deformation gradient tensor, $\bar{\mathbf{F}}$, is multiplicatively decomposed into an elastic (subscript $e$) and a plastic (subscript $p$) deformation gradient tensor: $\bar{\mathbf{F}}=\mathbf{F}_e\cdot\mathbf{F}_p$.

The following strain energy density is employed:

\begin{equation}
W=\frac{E(I_e-3-2\text{ln}(J_e))}{4(1+\nu)}+\frac{E\nu(\text{ln}(J_e))^2}{2(1+\nu)(1-2\nu)},
\end{equation}

\noindent where $E$ and $\nu$ denote Young's modulus and Poisson's ratio, respectively. Furthermore: $I_e=\text{tr}(\mathbf{F}_e^T\cdot\mathbf{F}_e)$ and $J_e=\text{det}(\mathbf{F}_e)$, where superscript $T$ denotes the transpose. Differentiating the strain energy with respect to $\mathbf{F}_e$ gives the $1^{\text{st}}$ Piola-Kirchhoff stress tensor, $\mathbf{P}_e$: $\mathbf{P}_e=\frac{\partial W}{\partial \mathbf{F}_e}$, which is related to the Mandel stress, $\mathbf{M}$, as $\mathbf{M}=\mathbf{F}_e^T\cdot\mathbf{P}_e$.

The employed yield function reads:

\begin{equation}
y=\sqrt{\frac{3}{2}\mathbf{M}^{dev}:\mathbf{M}^{dev}}-M_0-h\,\lambda^m,
\end{equation}

\noindent where material parameters $M_0$, $h$ and $m$ denote the initial yield stress, the hardening modulus and an exponential hardening parameter, respectively. Furthermore, $\mathbf{M}^{dev}$ denotes the deviatoric Mandel stress and $\lambda$ the plastic variable. The following associated flow rule is employed:

\begin{equation}
\dot{\mathbf{F}}_p=\dot{\lambda}\frac{\partial\,y}{\partial \mathbf{M}}\cdot\mathbf{F}_p.
\end{equation}

\noindent The Karush-Kuhn-Tucker conditions close the constitutive model:

\begin{equation}
y\leq 0, \quad\quad\quad\quad \dot{\lambda}\geq 0, \quad\quad\quad\quad \dot{\lambda}y=0.
\end{equation}

A periodic mesh is employed in the simulations, where the constraints due to the periodic boundary conditions are imposed using Lagrange multipliers. Dirichlet boundary conditions are used for the four corner nodes, where the displacement values are dictated by the right stretch tensor of the macroscale deformation $(\mathbf{U}^M$, which is symmetric). This results in the following system of linear equations, which must be constructed and solved for each iteration, for each increment:

\begin{equation}
\begin{split}
  \begin{bmatrix}\underline{\underline{K}}_\text{{int}}(\underline{u},\underline{z})& \left(\frac{\partial \underline{c}}{\partial \underline{u}}\right)^T\\\frac{\partial \underline{c}}{\partial \underline{u}} & \underline{\underline{0}}\end{bmatrix}\begin{bmatrix}d\underline{u}\\ d\underline{g}\end{bmatrix} \\ =\begin{bmatrix}\underline{f}_\text{{ext}}-\underline{f}_\text{{int}}(\underline{u},\underline{z})-\underline{g}^T\frac{\partial \underline{c}}{\partial \underline{u}}\\ \underline{c}(\underline{u})\end{bmatrix},
\label{DNS}
\end{split}
\end{equation}

\noindent where column $\underline{u}$ collects the displacement components of all nodes at an intermediate solution, column $\underline{z}$ the plastic variables in all quadrature points at an intermediate solution ($\lambda$ and $\mathbf{F}_p$), column $\underline{g}$ the Lagrange multipliers, column $\underline{c}$ the constraints due to the periodic boundary conditions, column $\underline{f}_\text{{ext}}$ the components of the reaction forces, column $\underline{f}_\text{{int}}$ the components of the internal forces  and matrix $\underline{\underline{K}}_\text{{int}}$ the derivatives of the internal force components with respect to the displacement components. $d\underline{u}$ and $d\underline{g}$ together denote the correction to the intermediate solution given by $\underline{u}$ and $\underline{g}$.

\section{POD-based model order reduction} \label{pod_explanation}

Projection-based MOR interpolates all $n_u$ kinematic variables, $\underline{u}$, using $n_b$ global basis functions according to:

\begin{equation}
\underline{u} \approx \sum_{i=1}^{n_b}\underline{\phi}_i\alpha_i=\underline{\underline{\Phi}}\,\underline{\alpha},
\end{equation}

\noindent where $\underline{\phi}_i$ of length $n_u$ denotes the $i^{\text{th}}$ basis function and scalar $\alpha_i$ denotes its associated weight that is to be computed online. $\underline{\underline{\Phi}}$ and $\underline{\alpha}$ collect all the basis functions and their associated weights in an $n_u \times n_b$ matrix and a column of length $n_b$, respectively.

In the POD method, global basis functions $\uubar{\Phi}$ are the orthonormal vectors corresponding to the largest singular values of an $n_u \times n_t$ matrix, $\underline{\underline{U}}$, storing $n_t$ training solutions. Alternatively, one can apply eigenvalue decomposition to product $\underline{\underline{U}}^T\underline{\underline{U}}$ (of size $n_t \times n_t$ ), resulting in an $n_t \times n_t$ matrix $\underline{\underline{V}}$ of which each column is an eigenvector of $\underline{\underline{U}}^T\underline{\underline{U}}$. The $n_t$ left-singular vectors (with the largest singular values) of $\underline{\underline{U}}$ are the columns of product $\underline{\underline{U}}\,\underline{\underline{V}}$, from which the first $n_b$ columns are utilized as the global basis functions.

In RVE simulations with periodic boundary conditions, the global basis functions are not directly extracted from the training solutions. Rather, the displacements are additively decomposed into a homogeneous contribution, $\bar{\underline{u}}$, and a fluctuating contribution, $\tilde{\underline{u}}$:

\begin{equation}
\underline{u}=\bar{\underline{u}}+\tilde{\underline{u}}.
\label{add_disp}
\end{equation}

Because the right stretch tensor of the macroscale deformation, $\mathbf{U}^M$, is the input for RVE simulations, $\bar{\underline{u}}$, can straightforwardly be computed. The global basis functions are thus only used to interpolate the fluctuating part of the kinematic variables, yielding the following expression:

\begin{equation}
\underline{u}=\underline{\underline{\Psi}}\,\underline{\omega}+\underline{\underline{\Phi}}\,\underline{\alpha},
\label{pod_split}
\end{equation}

\noindent where column $\underline{\omega}$ of length 3 is completely dictated by the known components of $\mathbf{U}^M$ (three components in 2D simulations) and matrix $\underline{\underline{\Psi}}$ (of size $n_u \times 3$ in 2D) homogeneously interpolates the displacement components.

According to Eqs.~(\ref{add_disp}) and~(\ref{pod_split}), the homogeneous deformations are first subtracted from the training solutions, and SVD is applied to the resultant (i.e.~fluctuating) displacement field.

The decomposition produces basis functions that are themselves periodic and hence, periodicity does not need to be actively enforced in the online simulations. Consequently, the system of linear equations that needs to be solved in terms of update $d\underline{\alpha}$ in the online simulations reads:

\begin{equation}
\begin{split}
\underline{\underline{\Phi}}^T\underline{\underline{K}}_\text{{int}}\underline{\underline{\Phi}}d\underline{\alpha} =  -\underline{\underline{\Phi}}^T\underline{\underline{f}}_\text{{int}} -\underline{\underline{\Phi}}^T\underline{\underline{K}}_\text{{int}}\underline{\underline{\Psi}}d\underline{\omega},
\end{split}
\label{POD_equation}
\end{equation}

\noindent where update $d\underline{\omega}$ is known.

\section{ANN-acceleration}

In this section, we discuss the RNN that rapidly emulates the basis coefficients in $\underline{\alpha}$ for each  increment (given by $\mathbf{U}^M$). This circumvents the iterative process (Eq.~(\ref{POD_equation})) necessary for conventional POD-based MOR.

\subsection{Network architecture}

\begin{figure}[htb!]
    \centering
    \includegraphics[width=0.42\textwidth]{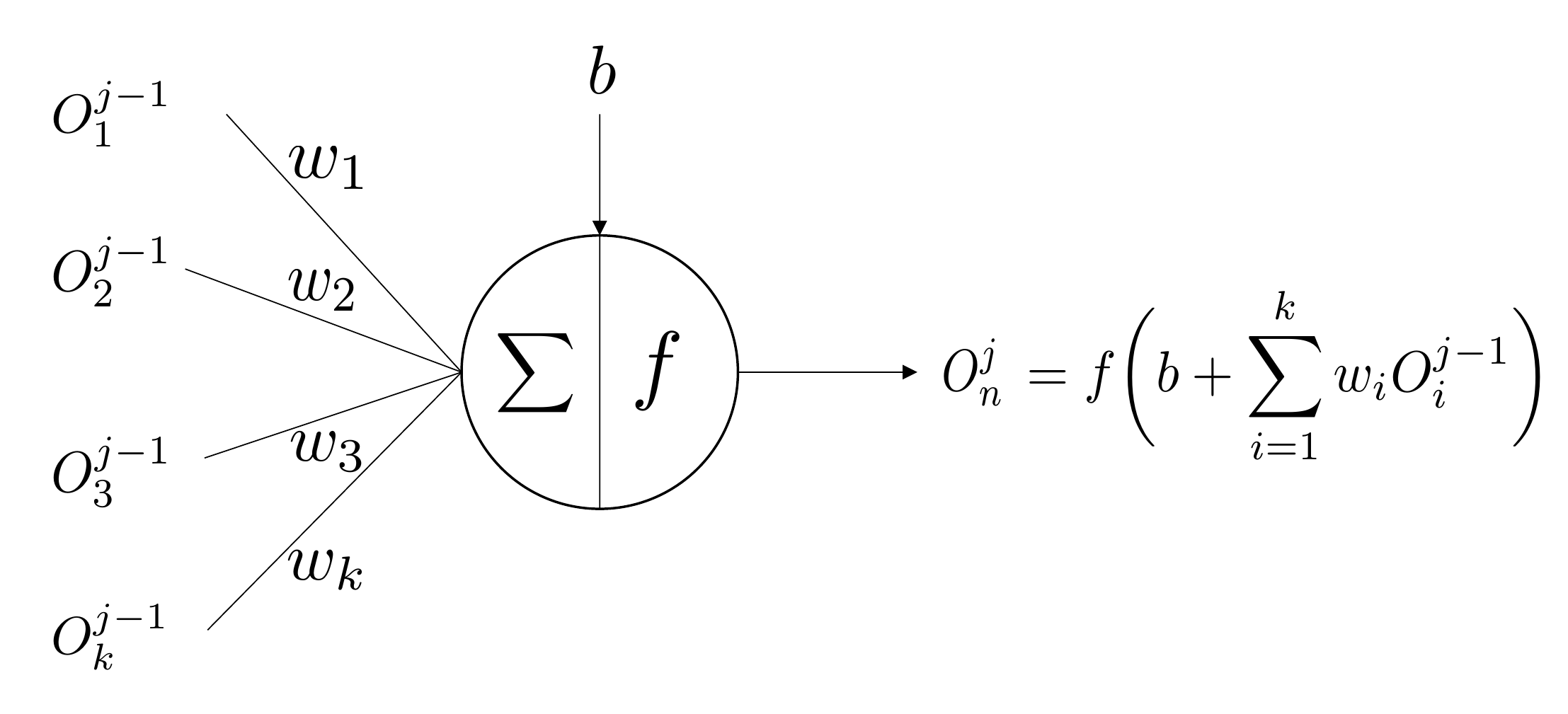}
    \caption{A single artificial neuron at layer $j$. The outputs of previous layer $O^{j-1}$ are the inputs of current layer $j$.}
    \label{a_neuron}
\end{figure}

\begin{figure}[htb!]
    \centering
    \includegraphics[width=0.42\textwidth]{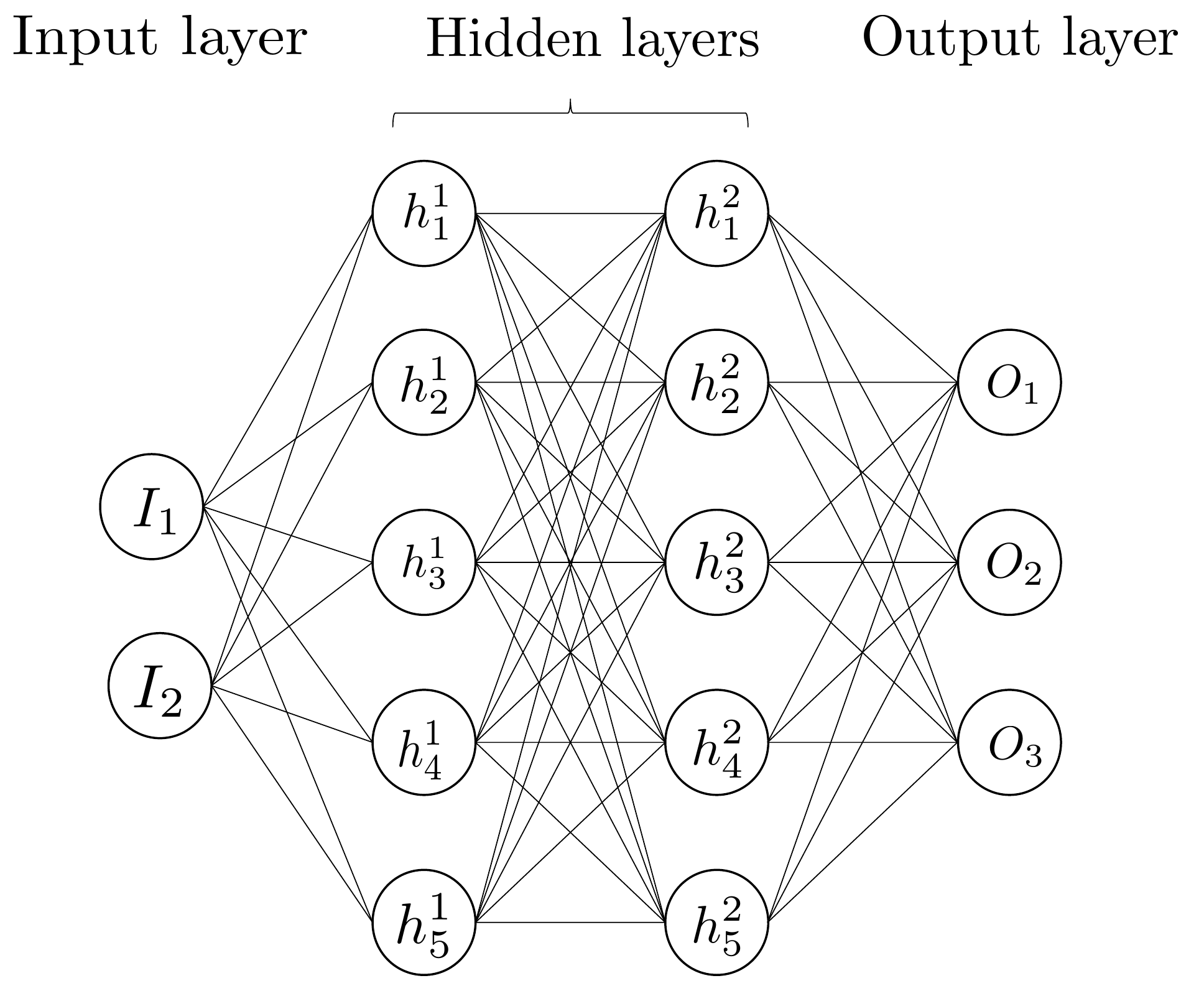}
    \caption{A feed forward neural network with three layers: Two hidden layers with five neurons each ($h_i^j$), two neurons for the input layer and three for the output layer.}
    \label{FFD_ANN}
\end{figure}

A neural network is a combination of numerous neurons. Each neuron receives several input values ($O^{j-1}_1$ to $O^{j-1}_k$ in Fig.~\ref{a_neuron}), and outputs a single value ($O^j_n$) as a function of weighing the input values ($w_i$), adding a bias to it ($b$), and inserting the result in an activation function ($f$). A collection of neurons are grouped together to form a layer. In a deep neural network, several layers of neurons are placed one after another (see Fig.~\ref{FFD_ANN}).

The best known ANNs are feed forward deep neural networks (see Fig.~\ref{FFD_ANN}). Feed forward networks have a unique relationship between input and output data, and cannot handle sequential information  (i.e.~an incremental sequence $\mathbf{U}^M$), as required for path-dependent models such as elastoplasticity \cite{ROCHA2020103995}.

\begin{figure}[htb!]
    \centering
    \includegraphics[width=0.50\textwidth]{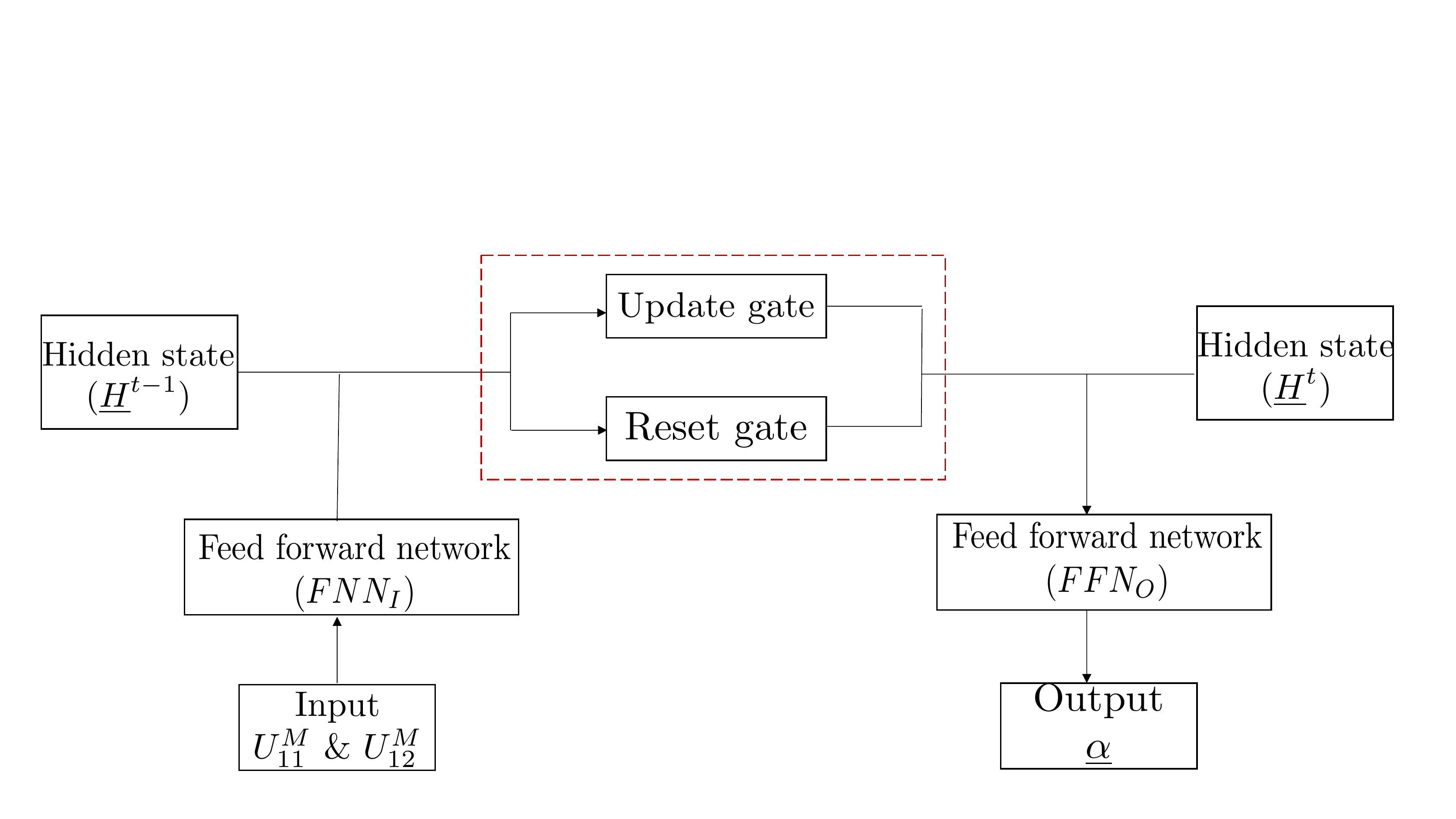}
    \caption{Neural network architecure used in this contribution. The red dashed box indicates the GRU.}
    \label{rnn_gru}
\end{figure}

On the other hand, recurrent neural networks (RNNs) have the intrinsic feature to handle sequential data. RNNs (Fig.~\ref{rnn_gru}) employ hidden state variables (`$H$') as memory elements to store and pass on information from the past to the upcoming sequence. Therefore, the predictions of an RNN at time step `$t$' are based on the input at time step (`$t$') and the values of the hidden variables (`$H_{t-1}$') at the beginning of that time step (`$t$'). Hidden variables `$H$' can be considered as the RNN\textquotesingle s way to quantify the past, in analogy to history variables $\underline{z}$ in the DNS and the conventional POD-based MOR.

Traditional RNNs suffer from the problem of vanishing gradients while handling long sequential data. This is due to the fact that the parameters at the beginning of the sequence depend on the gradient of the parameters present later in the sequence. In this process, the derivatives, which take small values, are multiplied several times, resulting in significantly smaller values, explaining the term vanishing gradients. To overcome this problem, we use an RNN with a Gated Recurrent Unit (GRU).

The GRU enables control over the flow of information through the hidden variables `$H$'. It uses an update gate and a reset gate to determine the amount of information to be passed on and to be retained by the hidden variable. The gated structure of a GRU also controls the flow of gradients during learning, such that the parameter update value does not vanish.

The RNN architecture used in this contribution is shown in Fig.~\ref{rnn_gru}. Adding Feed forward neural networks ($FFN_I$ and $FFN_O$) at the input and at the output of the GRU increases the accuracy of the RNN for complex problems \cite{WU2020113234}. 

\subsection{Learning phase of the recurrent neural network} \label{training_network}

We perform a supervised learning strategy in which the entire data is passed through the RNN numerous times, where every time the  data is passed through is referred to as an epoch. The learning stage, i.e. the identification of the weights and biases, is an iterative process in which the loss function (\ref{mse}) is minimized to increase the network\textquotesingle s accuracy. 

\begin{equation}
  L_{\text{MSE}} = \frac{1}{n} \sum_{i=1}^{n}\Arrowvert \underline{\alpha}_i-\underline{\alpha}_i^p \Arrowvert^2,
  \label{mse}
\end{equation}
\noindent where $n$ denotes the number of training solutions, superscript $p$ refers to the basis coefficients predicted by the RNN, and $\Arrowvert \bullet \Arrowvert$ denotes the $L^2$-norm.

An epoch has a forward propagation stage in which the information is passed from the input layer to the output layer. At the beginning of learning, the parameters such as weights and biases are initialized randomly.

In the backward propagation stage, the loss function\textquotesingle s gradients with respect to the RNN's parameters are calculated starting at the output layer and ending at the input layer. A gradient descent algorithm is used to update the RNN\textquotesingle s parameters. The procedure is followed for a number of epochs until the desired convergence is obtained.

The data is fed to the network in multiple batches in order to speed up the training process. Each batch consists of a sequence of input-output pairs that are extracted as training solutions at every load increment. The length of sequences is equal for all the batches. In the current contribution, data in each batch corresponds to the solutions of a single training simulation. There are 1000 load increments per simulation, therefore each batch is of length 1000. The hidden variables of the GRU unit are initialized as -1, as recommended in \cite{WU2020113234}.

\section{Results}

\subsection{Model setup and data collection}
\begin{figure}[htb!]
    \centering
    \includegraphics[width=0.25\textwidth]{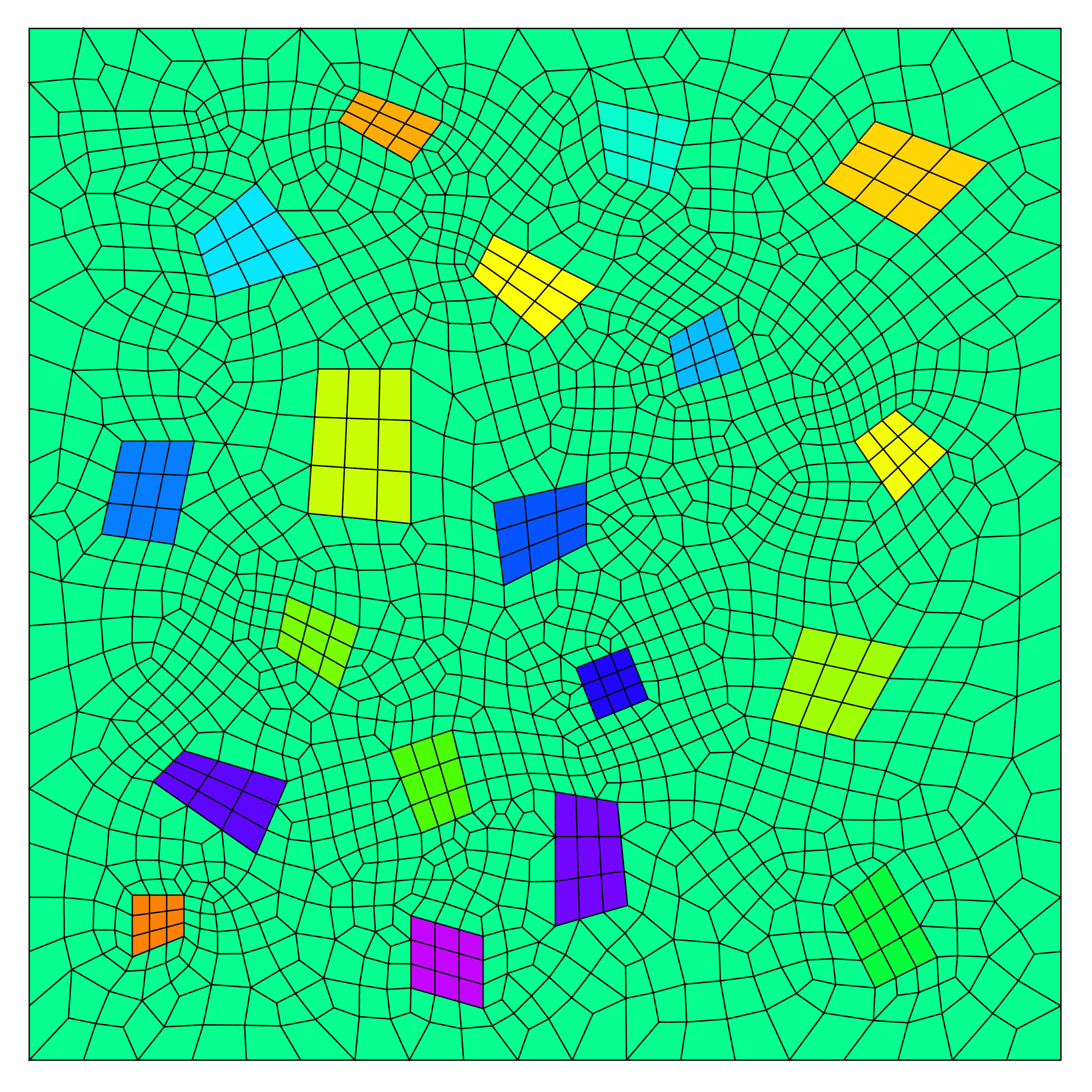}
    \caption{The discretized RVE with particles.}
    \label{rve}
\end{figure}

The discretized RVE is portrayed in Fig.~\ref{rve} and is subjected to cyclic and random loading. The material parameters for the matrix are set to $E=1$, $\nu=0.3$, $M_0=0.01$, $h=0.02$ and $m=1.05$. For the particles, the elastic material parameters are set to $E=20$, $\nu=0.3$, while $M_0=\infty$ ensures that the particles behave purely elastically. Because the matrix deforms mostly plastically and plastic deformation is isochoric, and because the particles deform only minimally relative to the matrix (due to the ratio of Young's moduli), we only consider the application of isochoric macroscale deformations (i.e.~$\textrm{det}(\mathbf{U}^M)=1$), governed by bounds $0.75<U^M_{11}<1.25$, $0.75<U^M_{22}<1.25$ and $-0.75<U^M_{12}<0.75$. This means that it is sufficient to only consider components $U^M_{11}$ and $U^M_{12}$ as input variables of the RNN (as $\textrm{det}(\mathbf{U}^M)=1$, $U^M_{11}$ and $U^M_{12}$ dictate the value of $U^M_{22}$).

\begin{figure}[htb!]
    \centering
    \includegraphics[width=0.50\textwidth]{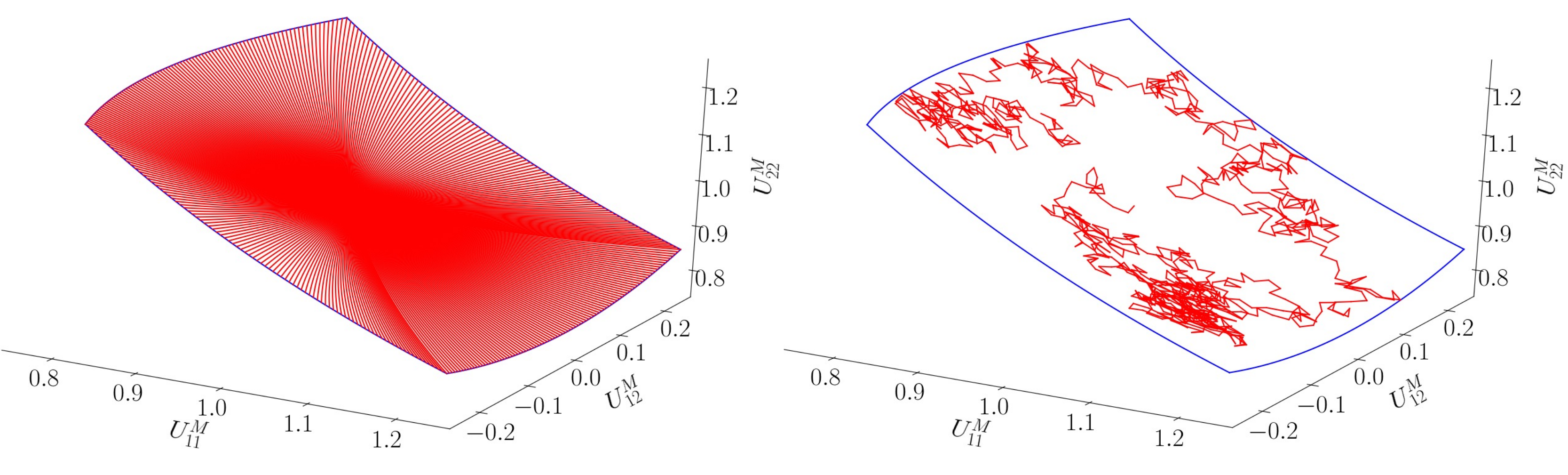}
    \caption{Left: Each red line presents the load path of a cyclic training simulation. Right: Load path of a single training simulation for random loading. Bounds $0.75<U^M_{11}<1.25$, $0.75<U^M_{22}<1.25$ and $-0.75<U^M_{12}<0.75$ of surface $\textrm{det}(\mathbf{U}^M)=1$ are presented by blue lines.}
    \label{training}
\end{figure}

A single RNN is simultaneously trained to emulate the basis coefficients for both cyclic and random loading. 350 cyclic loading training simulations (+10 validation simulations) and 10,000 random loading simulations (+100 validation simulations) are performed to determine the basis functions and to train the RNN. Random loading is not per se simulated because it is expected in nested multiscale simulations. Instead, it is considered to enhance the training because in true multiscale simulations with cyclic loading, the cyclic loading path of each RVE will slightly differ for each cycle \cite{WU2020113234}. The loading paths of the cyclic training simulations are presented in the left diagram of Fig.~\ref{training}. Each involves a loading stage and an unloading stage, each stage consisting of 500 increments. In the random loading simulations (one loading path is shown on the right in Fig.~\ref{training}), the loading direction is randomly selected for each of the 1000 increments and the loading step is fixed.

The RNN\textquotesingle s output variables are the coefficients of the POD basis, $\underline{\alpha}$. In order to obtain the input and output pairs for training the RNN, the POD problem described in section~\ref{pod_explanation} is solved for the same loading paths used for DNS. In this contribution 100 POD basis functions are used, whose coefficients are extracted at every load increment of each training simulation. Since elastoplasticity yields non-ellipticity, a large number of basis functions are required to obtain an acceptable accuracy.

\subsection{RNN predictions}
The learning stage of a neural network (i.e.~the algorithm to minimise the loss function in order to identify the network\textquotesingle s weights and biases) requires the selection of several hyperparameters. The selection of these hyperparameters affects the speed of the learning stage and the accuracy of the resulting network. In this contribution, the learning rate is set to 0.001. The mini-batch size is set to 1000. The parameters for ADAM optimizer are $\beta_1=0.9$, $\beta_2=0.999$  and $\epsilon=1e^{-8}$.

\begin{figure}[htb!]
   \centering
    \includegraphics[width=8.0cm,height=20cm,keepaspectratio]{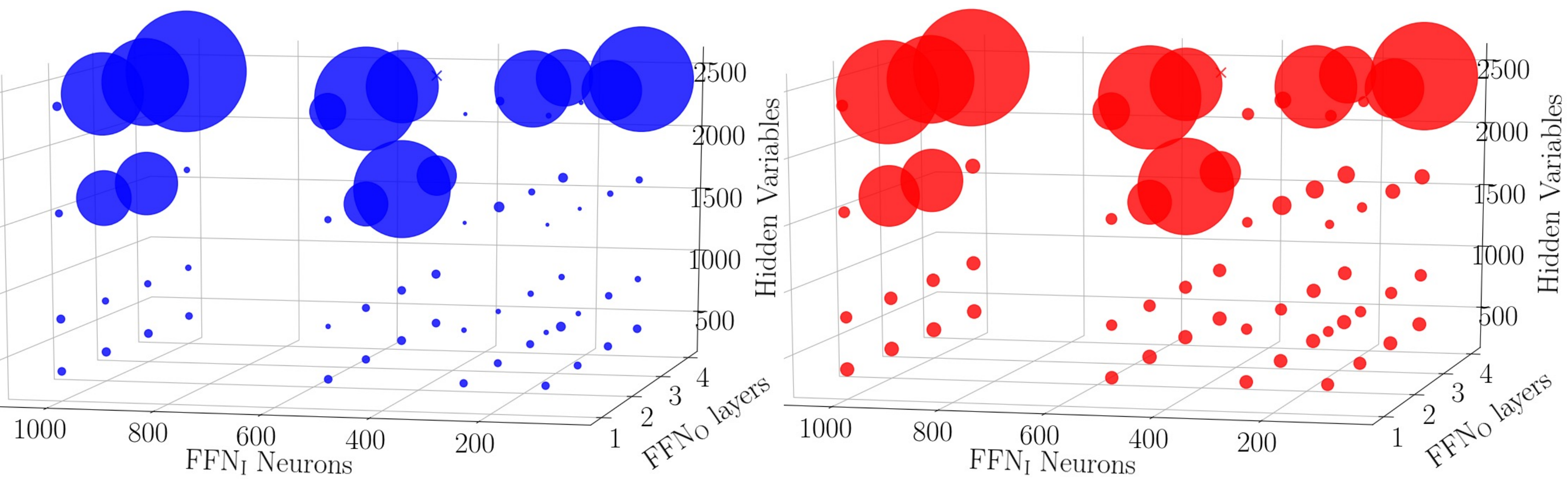}
      \caption{The loss function after training for 60,000 epochs for three neural network parameters: (i) Number of hidden layers in the $FFN_O$, (ii) Number of neurons in the hidden layers of $FFN_I$ and (iii) Number of hidden variables `$H$' of the GRU. The size of the circle corresponds to the value of the loss function (i.e.~a large circle corresponds to a large value of the loss function). Left: Loss function values for training data. Right: Loss function values for validation data.}
    \label{loss_fn}
\end{figure}

Different combinations of the numbers of layers, neurons and hidden variables are investigated with respect to the convergence of the loss function. Fig.~\ref{loss_fn} shows that the number of hidden variables and the number of hidden layers of $FFN_O$ influence the RNN\textquotesingle s accuracy the most.

Therefore, in this contribution the number of hidden layers in $FFN_I$, $FFN_O$ and the number of GRU layers is set to 1. The number of neurons in the hidden layer of the $FFN_I$ is set to 100. Both FFNs use the `Leaky ReLu' activation function. The number of hidden variables in the GRU and the number of neurons in the hidden layer of $FFN_O$ are set to 1600.

The RNN is trained for a total of 450,000 epochs, after which the loss function did not decrease further for both training and validation data. Each epoch consists of approximately $1\%$ of the whole training data in a mini-batch. Therefore, mini-batch is switched every 50 epochs to include all the training simulations. After training, the loss function (Eq.~(\ref{mse})) is reduced to $4.7 \cdot 10^{-5}$.

\begin{figure}[htb!]
    \centering
    \includegraphics[width=8cm,height=20cm,keepaspectratio]{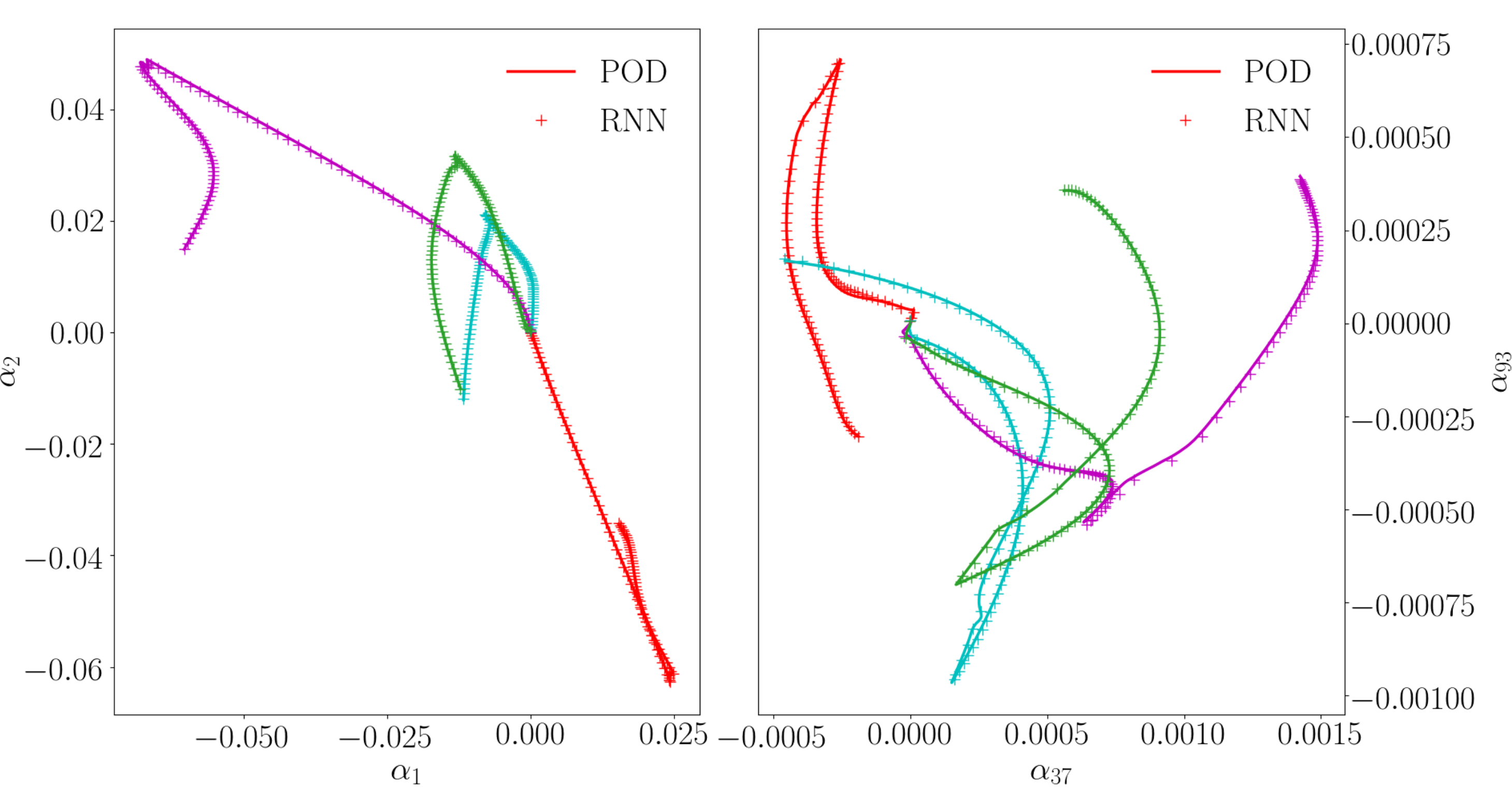}
    \caption{Cyclic loading validation simulations: Some RNN predictions (crosses) and the actual values (lines). The colors distinguish the four validation simulations.}
    \label{cyclic_rnn_a2}
\end{figure}

\begin{figure}[htb!]
    \centering
    \includegraphics[width=8cm,height=15cm,keepaspectratio]{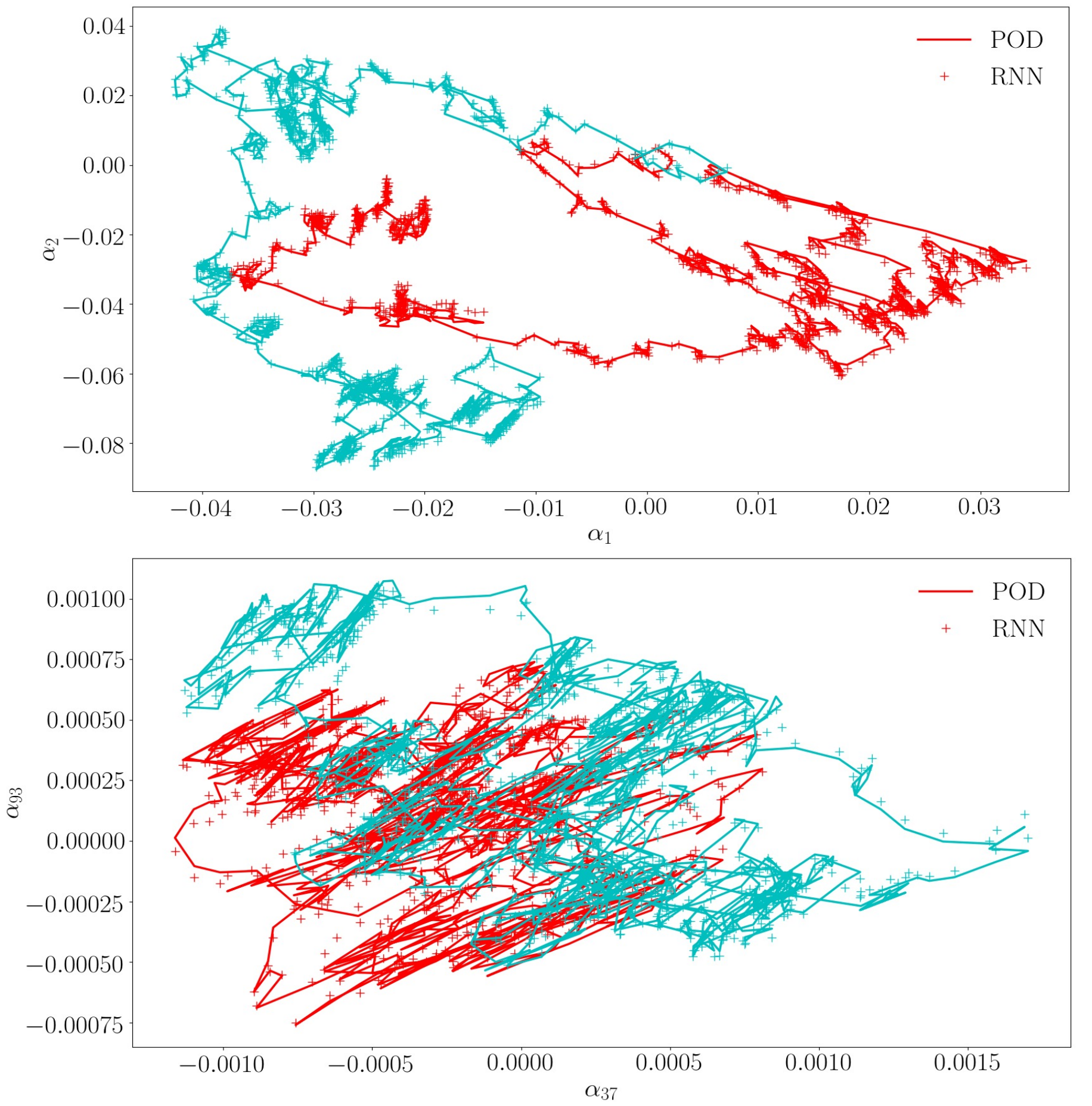}
    \caption{Random loading validation simulations: Some RNN predictions (crosses) and the actual values (lines). The colors distinguish two validation simulations.}
    \label{random_rnn_a2}
\end{figure}

Coefficients for some POD basis functions predicted by the RNN are presented in Figs.~\ref{cyclic_rnn_a2} (cyclic loading) and \ref{random_rnn_a2} (random loading), together with the exact coefficients. It is clearly visible that the RNN\textquotesingle s accuracy for cyclic loading is higher than for random loading, although the accuracy for random loading is still acceptable in our opinion. The average error calculated using Eq.~(\ref{mse}) for the 10 cyclic loading validation simulations is around $3\cdot 10^{-5}$, whereas the average error for the 100 random loading validation simulations is around $4\cdot 10^{-4}$.

\subsection{Mechanical predictions}

\begin{figure}[htb!]
    \centering
    \includegraphics[width=8cm,height=15cm,keepaspectratio]{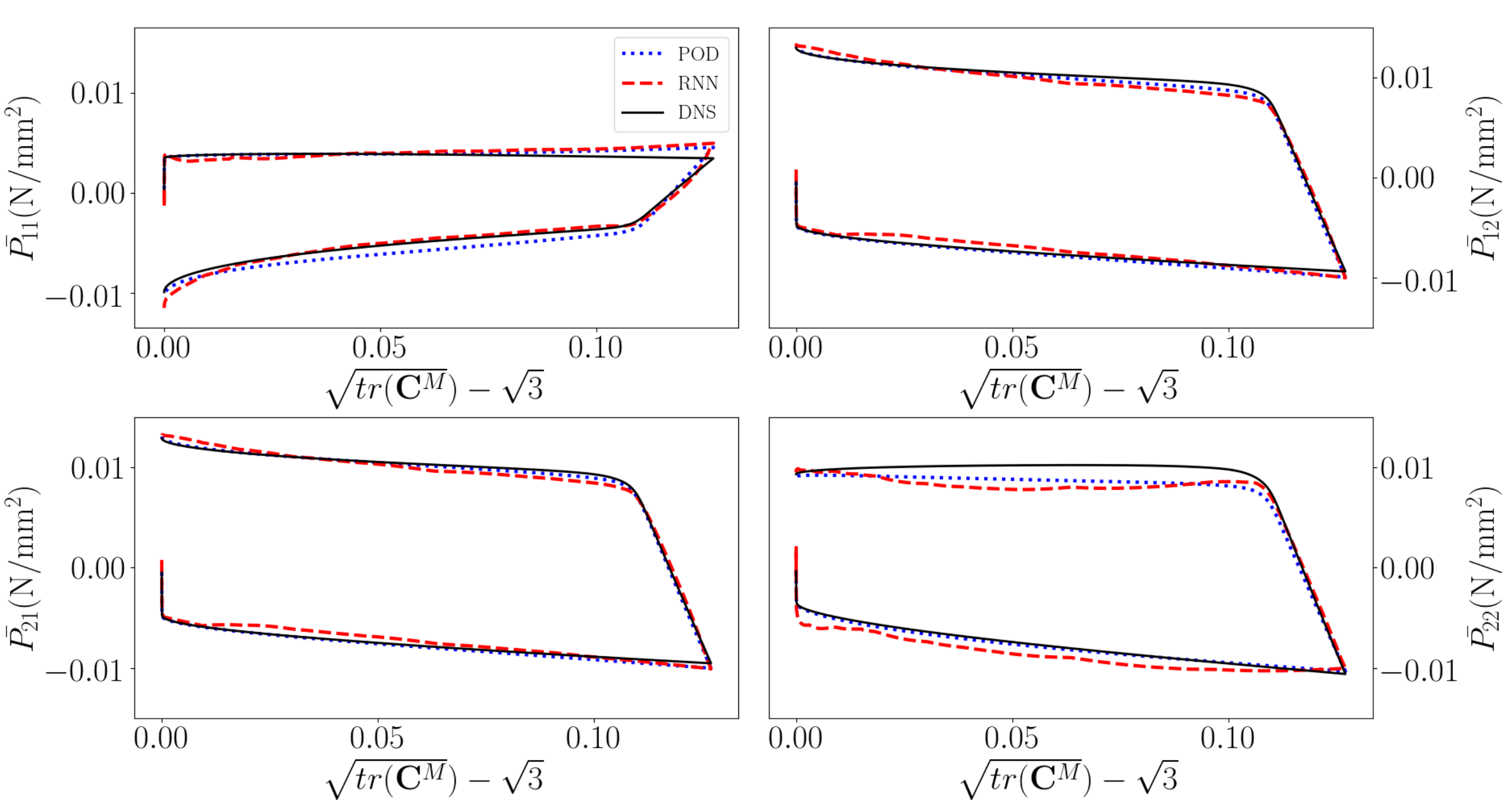}
    \caption{Components of the macroscale $1^{st}$ Piola-Kirchhoff stress as functions of the deformation for a cyclic loading validation simulation predicted by the DNS (black solid), by the conventional MOR (blue dashed), and by the RNN-accelerated MOR (red dotted).}
    \label{cyclic_pod}
\end{figure}

In this subsection, we compare the results of RNN-accelerated MOR with those of the conventional MOR and the DNS. We start with Fig.~\ref{cyclic_pod} in which the components of the homogenized $1^{\textrm{}st}$ Piola-Kirchhoff stress are presented for one of the cyclic loading validation simulations. We can see that the POD results match those of the DNS fairly accurately (albeit not perfectly), indicating that the number of 100 basis functions is sufficiently large. The results of the RNN-accelerated MOR also match those of the DNS and those of the MOR fairly accurately. Clearly, some differences are present, but the results indicate that the errors introduced by the RNN hardly influence the predicted macroscale stress.

\begin{figure}[htb!]
    \centering
    \includegraphics[width=8cm,height=20cm,keepaspectratio]{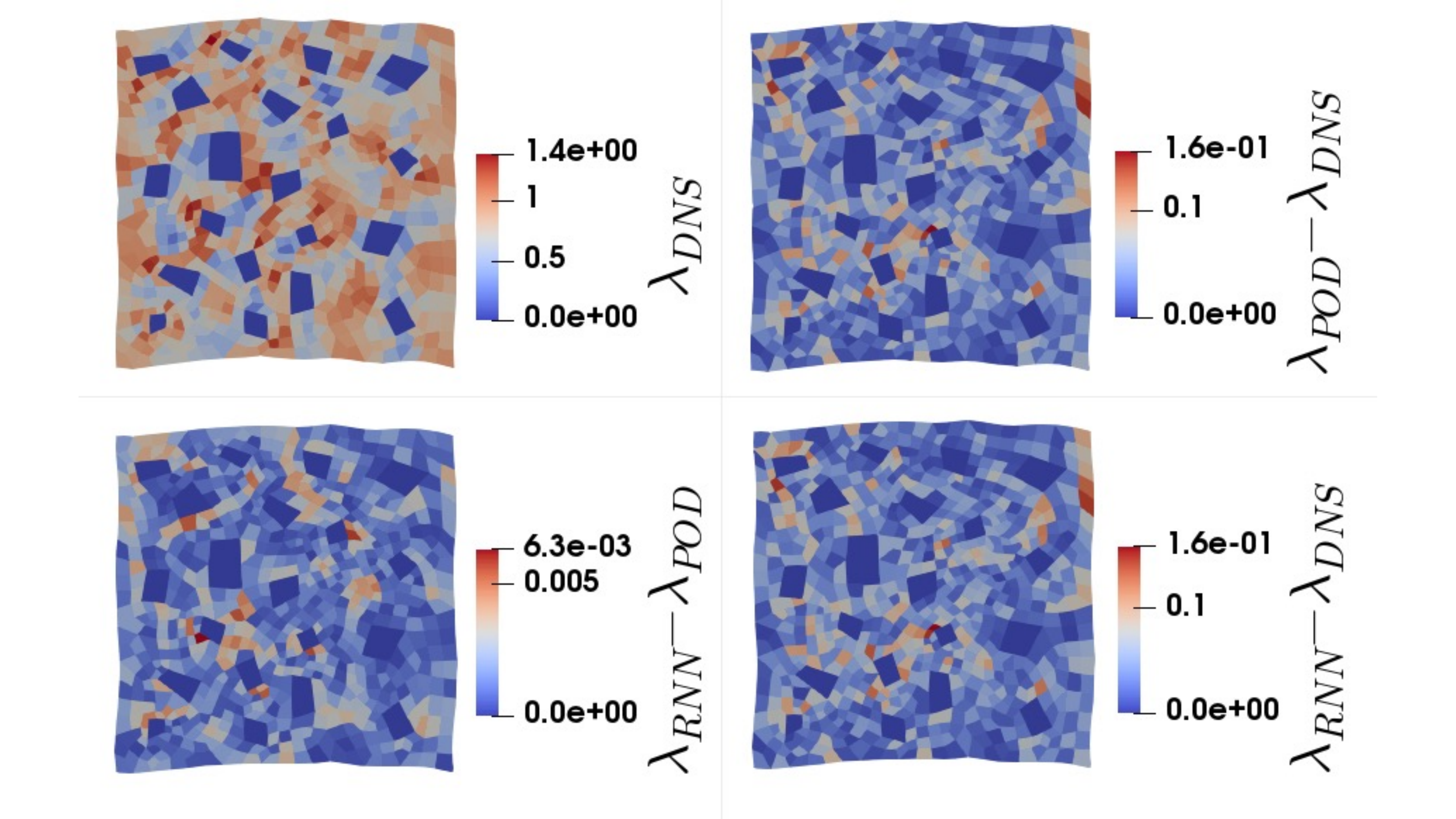}
    \caption{The plastic variable ($\lambda$) computed by the three methods for one of the cyclic loading validation simulations. Top-left: the DNS results, top-right: the difference between the POD results and the DNS results, bottom-left: the difference between the RNN-POD results and the POD results, bottom-right; the difference between the RNN-POD results and the DNS results.}
    \label{cyclic_epcum}
\end{figure}

As the appeal of RNN-accelerated MOR is the preservation of detailed information (microstructural information in case of RVEs), we also compare the plastic variables predicted by the different approaches. Fig.~\ref{cyclic_epcum} shows that the difference in the plastic variables predicted by the DNS and predicted by the MOR is not negligible, although also not completely unacceptable. In turn, the difference in the plastic variables predicted by the conventional MOR and predicted by the RNN-accelerated MOR is substantially smaller.

\begin{figure}[htb!]
   \centering
   \includegraphics[width=8cm,height=20cm,keepaspectratio]{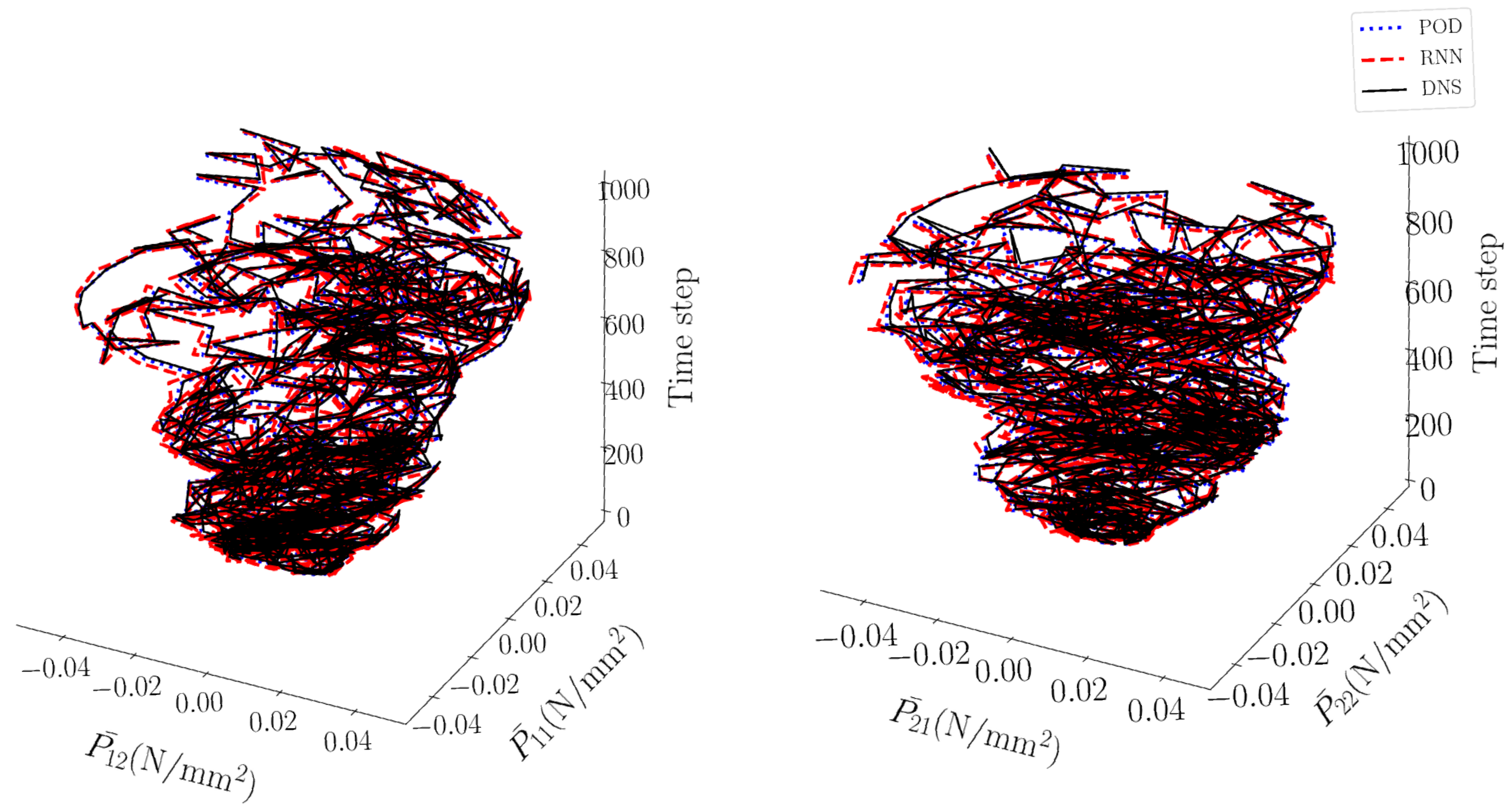}
   \caption{Components of the macroscale $1^{st}$ Piola-Kirchhoff stress values as functions of the number of increments for a random loading validation simulation predicted by the DNS (black solid), by the conventional MOR (blue dashed), and by the RNN-accelerated MOR (red dotted).}
    \label{random_PK}
\end{figure}

\begin{figure}[htb!]
    \centering
    \includegraphics[width=8cm,height=20cm,keepaspectratio]{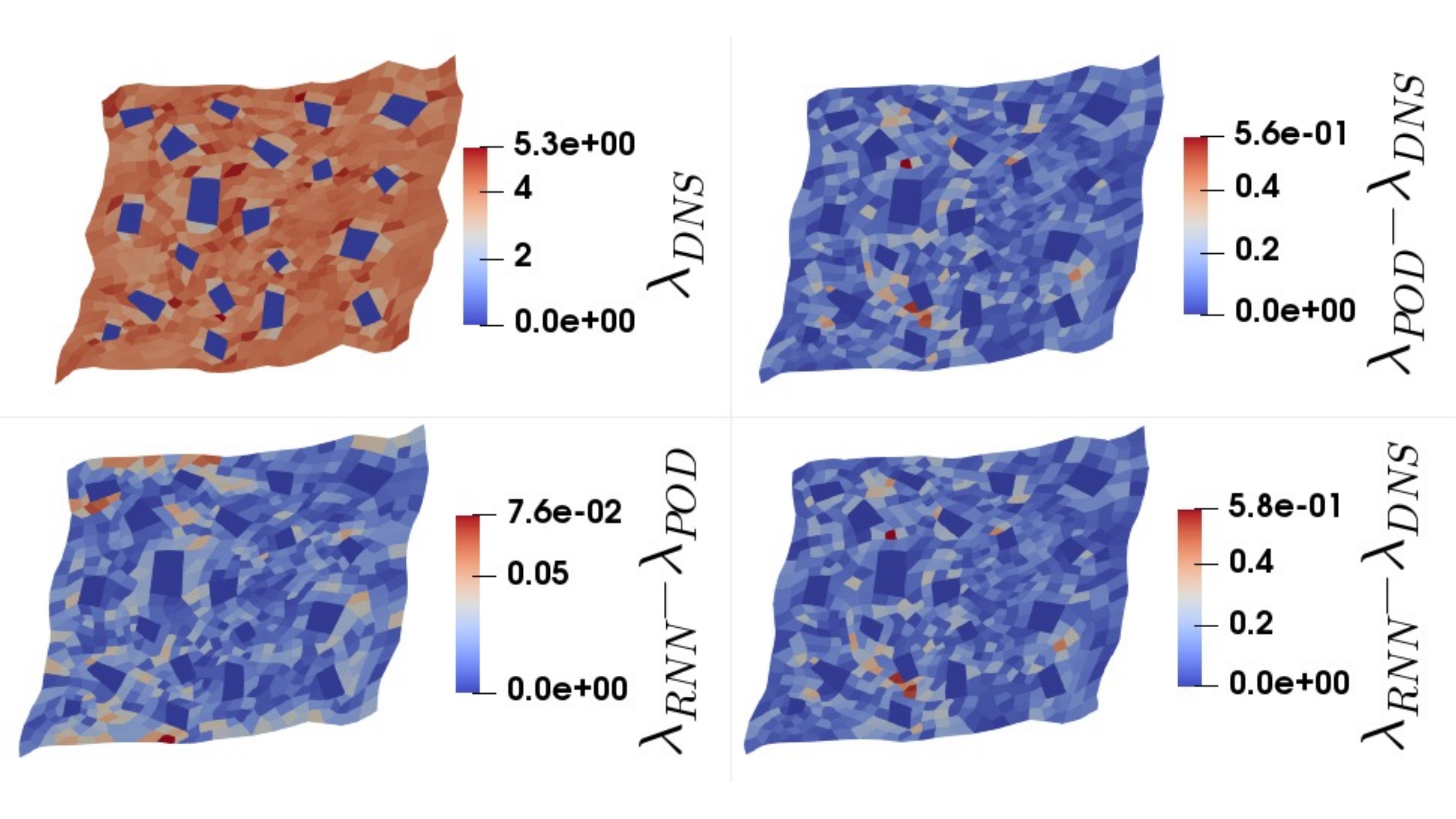}
    \caption{The plastic variable ($\lambda$) computed by the three methods for one of the random loading validation simulations. Top-left: the DNS results, top-right: the difference between the POD results and the DNS results, bottom-left: the difference between the RNN-POD results and the POD results, bottom-right; the difference between the RNN-POD results and the DNS results.}
    \label{random_epcum}
\end{figure}

We continue with results for random loading scenarios. In Fig.~\ref{random_PK}, the components of the $1^\textrm{st}$ Piola-Kirchhoff stress are again presented, but now for one of the random loading validation simulations and as a function of the increment number (instead of the deformation). On the other hand, Fig.~\ref{random_epcum} shows the plastic variables predicted by the different approaches. Comparing Fig.~\ref{cyclic_pod} with Fig.~\ref{random_PK} and Fig.~\ref{cyclic_epcum} with Fig.~\ref{random_epcum}, it can be concluded that the RNN-accelerated MOR is more accurate for cyclic loading than for random loading. The results can be argued to be sufficiently accurate, because, the random loading simulations were considered only to effectively train the RNN. But, in practice, the purpose of RNN accelerated MOR is to be utilized for loading cases arising in multi-scale simulations, which are closer to cyclic loading simulations.

The computational time required to prepare the training data and to test the validations are summarized in Table~\ref{cost_breakdown_offline} and \ref{cost_breakdown_online}. The training of the RNN was performed on HPC using 32GB of GPU computational resource for 7 days. Though the data preparation and training the RNN required a total of two weeks of computational time, the RNN-accelerated MOR is approximately 100 times as fast as the DNS and 22 times as fast as the conventional MOR in case of random loading. For cyclic loading on the other hand, the RNN-accelerated  MOR is only 13 times as fast as the DNS, whilst the conventional MOR is hardly faster than the DNS.

The difference in time savings of the RNN-accelerated MOR for cyclic and random loading are because the DNS and conventional MOR require more iterations for random loading than for cyclic loading. The reasons are that (1) the loading paths for cyclic loading are substantially shorter than those for random loading whilst the same number of increments is employed, and (2)  the previous plastic state is assumed at the start of each increment of a cyclic loading simulation (i.e.~DNS and conventional MOR) in order to increase the speed of the cyclic loading simulations.

\begin{table}[htb!]
  \scriptsize
  \begin{tabularx}{\columnwidth}{ X|X|X }
   \hline
    Data preparation & POD & RNN-POD \\ 
    \hline
    \hline
    Cyclic loading & 350$\times$1hr &350$\times$1hr\\
    & & \hspace{0.3cm} + \\
    & &350$\times$0.75hr \\
    \hline
      Random loading  & 10000$\times$7hr &10000$\times$7hr \\
    &  &\hspace{0.3cm} + \\
    & & 10000$\times$1.5hr \\
     \hline
  \end{tabularx}
  \caption{Computational time for data preparation}
  \label{cost_breakdown_offline}
\end{table}

\begin{table}[htb!]
  \scriptsize
  \begin{tabularx}{\columnwidth}{ X|X|X|X }
  \hline
   Online stage & DNS  & POD & RNN-POD \\
   \hline
   Cyclic loading & 55 min & 50 min & 4 min \\
   \hline
   Random loading & 7 hr & 1.5 hr & 4 min \\
   \hline 
  \end{tabularx}
  \caption{Computational time for validation simulations}
  \label{cost_breakdown_online}
\end{table}

%
%

\section{Conclusion}
In this contribution, a recurrent neural network (RNN) is used to emulate the basis function coefficients of projection-based model-order-reduction (MOR) for a representative volume element described by finite plasticity, subjected to cyclic loading and random loading. The RNN is simultaneously trained for cyclic loading and random loading. We have used an RNN, because elastoplasticity is history-dependent and  in analogy to the plastic variables in elastoplasticity, an RNN uses hidden variables to quantify its history.

Our results have shown that the RNN-accelerated MOR yields speed ups between factors 13 and 100 relative to the direct numerical simulations (and between factors 13 and 22 relative to conventional MOR). The accuracy is similar to conventional MOR, which is not entirely negligible relative to the direct numerical simulations. Nevertheless, with speeds up of up to factors of 100, the RNN-acceleration of MOR seems to make MOR for finite plasticity an interesting possibility - and perhaps also for other non-elliptical problems.


\section*{Acknowledgement}

S.~Vijayaraghavan, L.A.A.~Beex and S.P.A.~Bordas gratefully acknowledge the financial support of the Fonds National de la Recherche Luxembourg (FNR) grant INTER/FNRS/15/11019432/EnLightenIt/Bordas. This project has received funding from the H2020-EU.1.2.1.-FET Open Programme project MOAMMM under grant No 862015 and the EU\textquotesingle s H2020 project DRIVEN under grant No 811099. Computational resources have been provided by the supercomputing facilities C\'ECI funded by FRS-FNRS, Belgium and by the HPC of the University of Luxembourg~\cite{VBCG_HPCS14}.


Computational resources have been provided by the supercomputing facilities C\'ECI funded by FRS-FNRS, Belgium and by the HPC of the University of Luxembourg~\cite{VBCG_HPCS14}.


\bibliographystyle{abbrv}

\bibliography{pod_ann_bib}



\end{document}